%% file: root.tex
\documentclass[letterpaper, 10 pt, conference]{ieeeconf}  

\usepackage{amsmath} 
 
  {
      \newtheorem{assumption}{Assumption}
      
  }

\IEEEoverridecommandlockouts                              

\usepackage{graphicx} 
\usepackage{float}
\usepackage{relsize}
\usepackage{hyperref}
\usepackage{times} 

\usepackage{amssymb}  
\usepackage{caption}
\usepackage{subcaption}
\usepackage[font=small]{caption}
\usepackage{todonotes}[inline]  
\usepackage[normalem]{ulem}

\title{\LARGE \bf
In Situ Translational Hand-Eye Calibration \\ of  Laser Profile Sensors using Arbitrary  Objects
}

\author{Prajval Kumar Murali$^{*}$, Ines Sorrentino$^{*}$, Angelo Rendiniello, Claudio Fantacci, Enrico Villagrossi,  \\
Andrea Polo, Alessandro Ardesi, Marco  Maggiali, Lorenzo Natale, Daniele Pucci  and Silvio Traversaro
\thanks{$^{*}$ The authors contributed equally to this article}%
\thanks{I. Sorrentino, A. Rendiniello, M. Maggiali, L. Natale, D. Pucci, S. Traversaro are with the Fondazione Istituto Italiano di Tecnologia, Via San Quirico 19D, 16163, Genova, Italy (e-mail: name.surname@iit.it)}%
\thanks{C. Fantacci is with the Fondazione Istituto Italiano di Tecnologia, Via San Quirico 19D, 16163, Genova, Italy (now at DeepMind, e-mail: name.surname@gmail.com).}%
\thanks{P.K. Murali is with the Fondazione Istituto Italiano di Tecnologia, Via San Quirico 19D, 16163, Genova, Italy (now at BMW Group, e-mail: prajval.murali@gmail.com).}%
\thanks{A. Polo, A. Ardesi are with Danieli Automation S.p.A., Via B. Stringher 4, 33042, Buttrio, Italy (e-mails: a.polo@dca.it, a.ardesi@dca.it)}%
\thanks{E. Villagrossi is with Danieli Automation S.p.A., Via B. Stringher 4, 33042, Buttrio, Italy (now at CNR-STIIMA e-mail: name.surname@gmail.com)}%
}

\begin{document}

\maketitle
\thispagestyle{empty}
\pagestyle{empty}

\begin{abstract}
Hand-eye calibration of laser profile sensors is the process of extracting the homogeneous transformation between the laser profile sensor frame and the end-effector frame of a robot in order to express the data extracted by the sensor in the robot's global coordinate system. For laser profile scanners this is a challenging procedure, as they provide data only in two dimensions and state-of-the-art calibration procedures require the use of specialised calibration targets. This paper presents a novel method to extract the translation-part of the hand-eye calibration matrix with rotation-part known \textit{a priori} in a \textit{target-agnostic} way. Our methodology is applicable to \textit{any} 2D image or 3D object as a calibration target and can also be performed \textit{in situ} in the 
final application. The method is experimentally validated on a real robot-sensor setup with 2D and 3D targets. 
\end{abstract}


\input{sections/introduction.tex}

\input{sections/background.tex}

\input{sections/methods.tex}

\input{sections/experiments.tex}

\input{sections/conclusions.tex}


\bibliography{IEEEexample}
\bibliographystyle{IEEEtran}

\end{document}

%% file: sections/introduction.tex
\section{Introduction}
\label{sec:introdcution}

\begin{figure}[t!]
     \centering
     \begin{subfigure}[b]{0.235\textwidth}
         \centering
         \includegraphics[width=\textwidth, height = 6cm]{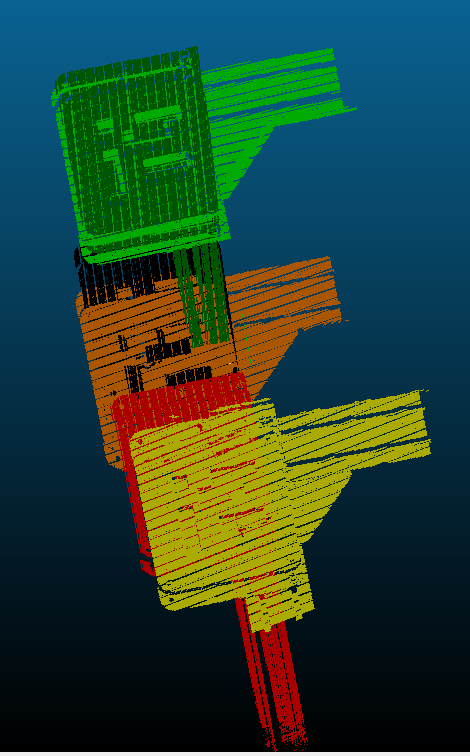}
         \caption{Incorrect hand-eye calibration}
         \label{fig:uncalibrated}
     \end{subfigure}
     \hfill
     \begin{subfigure}[b]{0.235\textwidth}
         \centering
         \includegraphics[width=\textwidth, height = 6cm]{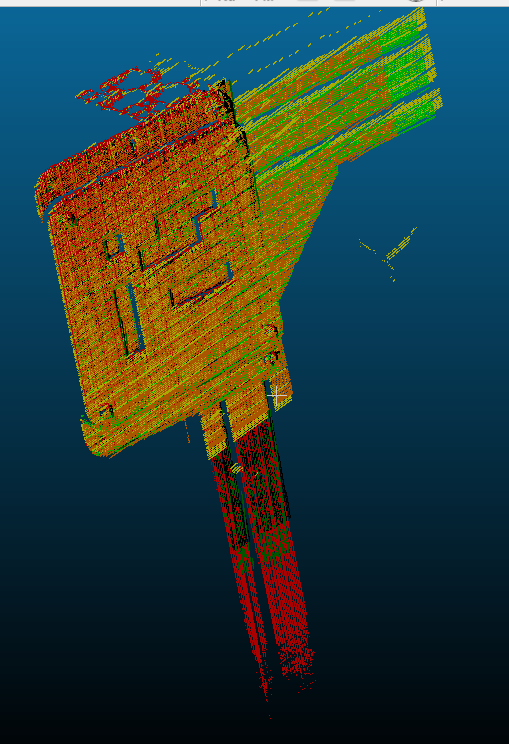}
         \caption{Correct hand-eye calibration}
         \label{fig:calibrated}
     \end{subfigure}
        \caption{The effect of incorrect translation (\ref{fig:uncalibrated}) and correct translation (\ref{fig:calibrated}) in the hand-eye calibration matrix ${}^E H_{S}$ of a stationary object with different scanning trajectories. The rotation part (${}^E R_{S}$) of ${}^E H_{S}$ is correct and identical in both cases.}
        \label{fig:caluncal}
\end{figure}

A laser profile sensor projects a laser line onto a surface and captures the 2D intensity laser profile in the sensor plane. In industrial settings, they have numerous applications including surface inspection, object pose detection, machining, model scanning. A laser profile sensor attached to the end-effector of a robot allows inspection at a range of locations feasible within the kinematic constraints of the robot. 
The 3D surface reconstruction is made by combining multiple measurements while moving the laser profile sensor over the surface of the target. In order to do this, it is necessary to merge the 2D measurements from the laser sensor with the robot pose to form the 3D spatial point cloud. There are two methods to form the 3D point cloud from 2D laser profile sensor data: \textit{stop and look} and \textit{buffered synchronization}~\cite{borangiu2010robot}. In the first method, the robot is moved in a step-wise manner such that the laser sensor is over the target object. A profile is not captured when the robot is in motion. For every profile the corresponding robot pose is saved. In the second method, the laser sensor captures data while the robot is in motion. A synchronization signal is used to match the laser data and robot poses~\cite{wagner2015self}. However in both cases, in order to reconstruct the point cloud in 3D, the homogeneous transformation between the sensor's coordinate frame and the coordinate frame of the robot end-effector must be known, the so called hand-eye calibration. If this transformation is not known or incorrect, then the data collected at multiple positions cannot be transformed into one global coordinate frame. The hand-eye calibration sources of errors, their analysis and solutions are presented in \cite{Zou2018, Zhang2020}. For illustration, Figure \ref{fig:caluncal} shows the effect of an incorrect translation-part of the hand-eye calibration (Figure \ref{fig:uncalibrated}) and correct calibration (Figure \ref{fig:calibrated}) for multiple acquisitions of a stationary target. 

Unlike typical camera-based vision sensors, there is no single widely adopted strategy for hand-eye calibration of 2D laser sensors because a single measurement from the sensor provides only 2D data~\cite{sharifzadeh2020robust}. 
In order to extract the hand-eye rotation and translation, a specially structured calibration artefact or target is usually necessary. In literature, various calibration artefacts have been used such as spherical balls (\cite{zhu2005scanner,li2011calibration,ren2012calibration,yin2014novel,che2000ball,liska2018hand,shen2012robotic}), planar disks (\cite{chen2016noise}), calibration pins (\cite{wagner2015self}), planar objects (\cite{sharifzadeh2020robust}), pyramids (\cite{antone2007fully}) or objects with multiple stepped surfaces \cite{santolaria2011crenellated}. Some of the calibration techniques require manual data collection. For instance in \cite{yin2013vision}, the practitioners are required to manually move the laser profile until it passes through a fixed X-shaped marker. 
The advantage of a simple artefact in this case is obscured by the requirement of highly specialised user skills to align and move the robot. On the other hand, the upside to having complex artefacts is that calibration can be performed autonomously with minimal user interaction during the data collection phase. However, increased complexity of the target artefacts result in higher difficulty to manufacture and increased cost. Furthermore, unavailable or inaccurate artefacts renders the hand-eye calibration impractical as all the approaches in literature are tightly coupled to the corresponding calibration artefact.
In some situations the orientation of the sensor is already known \textit{a priori} due to how the sensor is mounted on the robot~\cite{yang2019}. 

In this paper, we show how to relax the assumption of using a specific calibration artefact, and we demonstrate how to perform the translation calibration with any 2D or 3D target. To the best of our knowledge, no other work in literature presents a target-agnostic method to perform hand-eye translation calibration of a laser profile sensors.
The key contribution 
is a novel target-agnostic method to extract the translation-part of the hand-eye calibration that is applicable with any arbitrary 2D or 3D targets. In fact, looking to an industrial robotic application, the calibration target can be any object that can be scanned by the robot in the workcell, that's why we term as \textit{in situ} calibration. The only constraint is the knowledge of the Computer-aided design (CAD) model of the object used as a calibration target. 



%% file: sections/background.tex
\section{Background}
\label{sec:background}

\subsection{Mathematical Notation}\label{subsec:math_notation}

The notation used in the paper is detailed below.

\begin{itemize}
	
	\item Let $p \in \mathbb{R}^3$ denote a 3D column vector representing a point in $\mathbb{R}^3$.
	
	\item Let $\bar p \in \mathbb{R}^{4}$ denote the homogeneous representation of the vector $p$ defined as $\bar p = \begin{bmatrix} p & 1 \end{bmatrix}^T$.
	
	\item Let $\mathbf{P} = \{ \bar p_i \} _1^r$ denote a laser profile defined as a set of homogeneous vectors.
	
	
	\item Let $\mathbf{\Pi}$ denote a point cloud defined by a set of homogeneous vectors and reconstructed from a set of laser profiles.
	
	
	\item Let ${}^A o_B \in \mathbb{R}^3$ be the position of the frame B with respect to the frame A expressed in frame A.
	
	\item Let ${}^A R_B \in \mathbb{R}^{3 \times 3}$ be the rotation matrix describing the orientation of the frame B with respect to the frame A.
	
	\item Let ${}^A H_B \in \mathbb{R}^{4 \times 4}$ be the homogeneous transformation matrix describing the transformation of a point from the frame B to the frame A.
	
	\item Let $1_m \in \mathbb{R}^{m \times m}$ be the identity matrix of dimension $m \times m$
	
	
	\item Let $0_{m \times l} \in \mathbb{R}^{m \times l}$ be the null matrix of dimension $m \times l$.
	
	\item Let $\bigcup$ denote the union of a collection of sets that is the set of all elements in the collection. In this paper this is used to represent when two or more point clouds are merged together, after they have been expressed in the same frame.
\end{itemize}

\subsection{Robot-Sensor system}

\begin{figure}[t!]
	\centering
	\includegraphics[width = \columnwidth, height = 6cm]{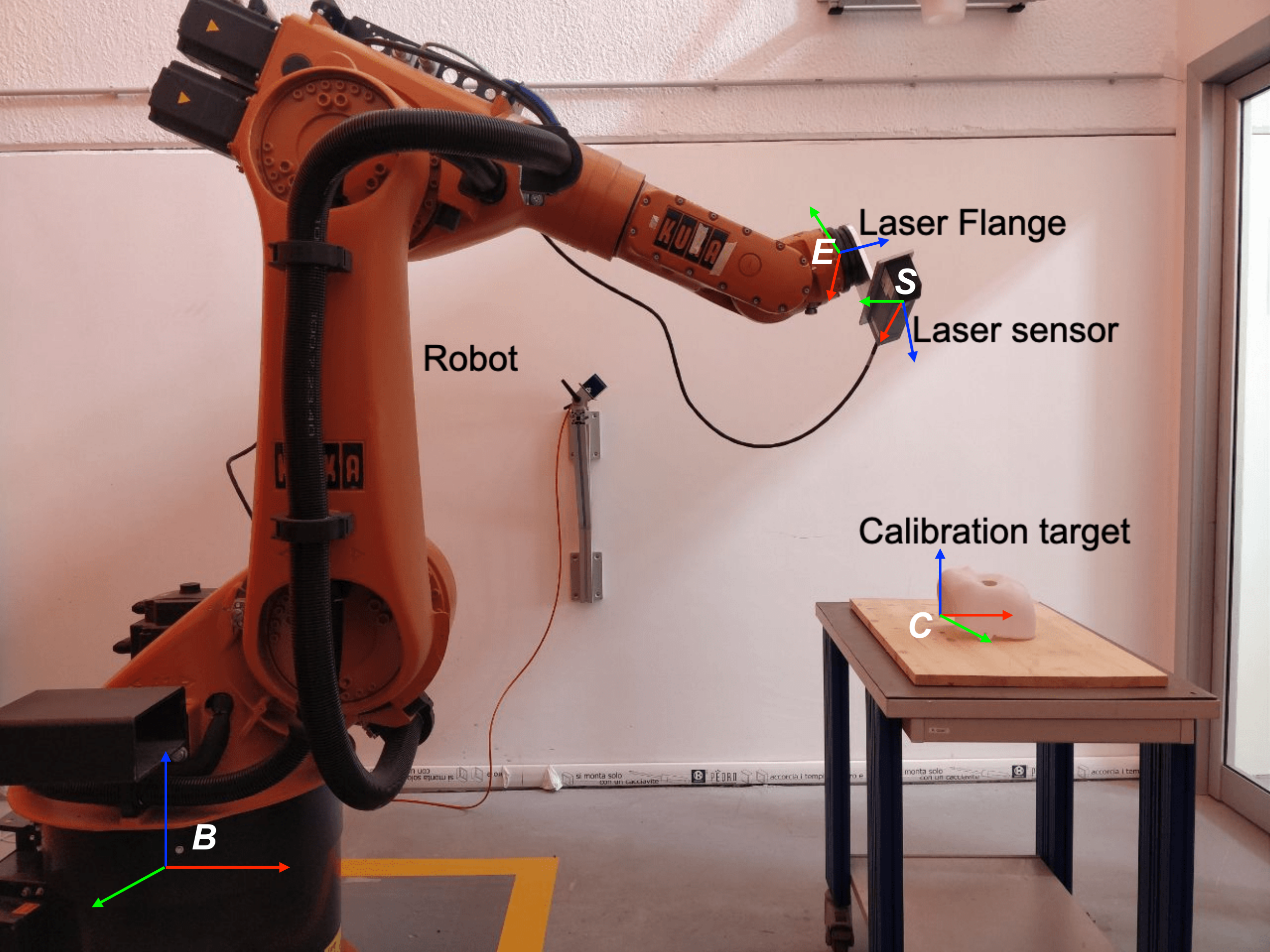}
	\caption{Robot-sensor system used as the experimental setup. The image illustrates the frames attached to the robot base ($B$), the robot end-effector ($E$), the sensor ($S$) and the target ($C$).
    }
	\label{fig:frames}
\end{figure}

A robot-sensor system is defined in terms of three coordinate systems: the robot base coordinate system $B$, the end-effector coordinate system $E$ and the sensor coordinate system $S$ as shown in Figure \ref{fig:frames}.
The objective of the hand-eye calibration, also called robot-sensor problem, is to find the homogeneous transformation ${}^E H_S$ that describes the position and the orientation of the sensor coordinate system $S$ with respect to the end-effector coordinate system $E$ \cite{siciliano2010robotics}.
In particular, the homogeneous matrix ${}^E H_S \in \mathbb{R}^{4 \times 4}$ is defined as:
\begin{equation}
{}^E H_S = \begin{bmatrix} {}^E R_S & {}^E o_S \\  0_{1\times 3} & 1 \end{bmatrix}  ,
\end{equation}
where ${}^E R_S \in \mathbb{R}^{3 \times 3}$ is the rotation matrix between sensor frame $S$ and the end-effector frame $E$ and ${}^E o_S \in \mathbb{R}^{3}$ is the translation of the origin of the sensor frame $S$ with respect to the origin of the end-effector frame $E$ expressed in the end-effector frame.

\subsection{Laser sensor and Point Cloud definition}

The class of sensor considered in this paper is laser profile sensor that measures the linear distance between the sensor and the objects that lay in its reception plane and the amount of light reflected by the objects. In particular, laser profile sensors projects a laser line on an object and provides a 2D profile in the sensor plane ${}^S P$. The acquired 2D profiles can be transformed from the sensor coordinate system $S$ into the 3D robot base frame $B$.

In the rest of the paper, we use the generalization of the operations that are typically defined for vectors. 
For example, if $q \in \mathbb{R}^3$ is a vector we can define the following relationship:
\begin{equation}     \mathbf{P} + \begin{bmatrix}q \\ 0\end{bmatrix} :=  
    \{ \bar p + \begin{bmatrix}q\\ 0\end{bmatrix}~|~\bar p \in  \mathbf{P} \}  ,
\end{equation}
while with ${}^A H_S {}^S \mathbf{\Pi}$ we indicate the laser profile obtained applying to  all its points the ${}^A H_S$ transform:
\begin{equation}
    {}^A H_S {}^S \mathbf{P} :=  
    \{ {}^A H_S {}^s \bar p~|~{}^s \bar p_i \in \mathbf{P} \} .
\end{equation}
A single 2D profile may not be sufficient to perform the localization or even just the detection of the object of interest. For this reason, typically the robot end-effector with the sensor mounted on it is moved in $n$ different positions and performs $n$ profile acquisitions $\mathbf{P}$. From the set of laser profiles it is possible to reconstruct a point cloud ${}^S \mathbf{\Pi}$. Indeed, if each profile ${}^{S_j} \mathbf{P}_j$ is transformed from the sensor frame $S_j$ in the $j$-{th} position to the robot base frame $B$ and the union of the set of profiles is performed, we obtain the point cloud ${}^B \mathbf{\Pi}$ of the scanned scene expressed in the frame B:
\begin{equation}\label{pointcloud}
{}^B \mathbf{\Pi} = \bigcup_{j = 1}^{n} {}^B H_{S_j} {}^{S_j} \mathbf{P}_j .
\end{equation}

In the rest of the paper, we use the terms \textit{profile}, \textit{reconstructed point cloud} and \textit{dataset} considering the following definitions. A \textit{profile} is the set of points corresponding to the laser line, acquired by the sensor in a given position. By moving the robot without changing the end effector orientation, we collect $n$ profiles and via the Equation \eqref{pointcloud} we can obtain a \textit{reconstructed point cloud}. For performing a laser sensor calibration we need to collect $m$ reconstructed point clouds with different end-effector orientations, and the collection of $m$ reconstructed point clouds represents a \textit{dataset}.

For the next section, it is useful to express the transformation matrix ${}^B H_{S_j}$  as two consecutive transformations:
\begin{equation}
{}^B H_{S_j} = {}^B H_{E_j} {}^{E_j} H_{S_j}  ,
\end{equation}
where ${}^B H_{E_j}$ is typically computed from the robot kinematic model and the robot joint positions \cite{siciliano2010robotics}. Furthermore, we can write ${}^{E_j} H_{S_j} = {}^{E} H_{S}$  that allows us to remove the dependency on the index $j$, as this transformation matrix does not change while the robot moves.
This permits to write equation~\ref{pointcloud} as:
\begin{equation}\label{pointcloudextended}
{}^B \mathbf{\Pi} = \bigcup_{j = 1}^{n} {}^B H_{E_j} {}^{E} H_S {}^{S_j} \mathbf{P}_j .
\end{equation}

%% file: sections/methods.tex
\section{Methods}
\label{sec:methods}

In this section, we present an algorithm to calibrate the hand-eye translation under the following assumptions.
\begin{assumption}
\label{ass:rotationIsKnown}
${}^E R_S$ is known a-priori, while ${}^E o_S$ is unknown and needs to be identified. 
\end{assumption}
\begin{assumption}
\label{ass:eeRotationIsConstant}
For the calibration process a single \textit{dataset} composed of multiple \textit{reconstructed point clouds} of a given object are acquired. Each \textit{reconstructed point cloud} is obtained by moving the end effector  of the robot ${}^B o_{E_i}$ and collecting \textit{profiles} while keeping the orientation of the robot end effector with respect to the base fixed, i.e., :
\begin{equation*}
    \forall~i \hspace{1em} {}^B R_{E_i} = {}^B R_{E_1} .
\end{equation*}
\end{assumption}
\begin{assumption}
\label{ass:endeffectorRotationIsKnown}
For each profile acquisition, the pose of the end effector of the robot with respect to base ${}^B R_{E_i}$ is known as it is computed using the joint positions measurements and the geometric model of the robot. 
\end{assumption}


Under Assumption~\ref{ass:rotationIsKnown} we can define a frame $E[S]$ that is oriented as the sensor frame, but whose origin is centered in the end-effector frame $E$, i.e.,:

\begin{equation}
\label{eq:frame_with_only_rotation_known}
    {}^E H_{E[S]} = 
    \begin{bmatrix}
    {}^E R_{S}  &  0_{3 \times 1} \\
    0_{1 \times 3}  & 1
    \end{bmatrix}
    .
\end{equation}

If the translation of the sensor frame ${}^E o_S$ is not known, it is not possible to reconstruct a \textit{reconstructed point cloud} from a set of profiles using equation~\ref{pointcloudextended}. However, we can reconstruct a related point cloud ${}^B \mathbf{\Pi}'$ by using ${}^E H_{E[S]}$ in place of ${}^E H_{S}$ in Equation~\ref{pointcloudextended}, only using quantities that are known according to Assumptions~\ref{ass:rotationIsKnown} and \ref{ass:endeffectorRotationIsKnown}:
\begin{equation}\label{pointcloudcorrupted}
{}^B \mathbf{\Pi}' = \bigcup_{j = 1}^{n} {}^B H_{E_j} {}^{E} H_{E[S]} {}^{S_j} \mathbf{P}_j  .
\end{equation}

To highlight how ${}^B \mathbf{\Pi}'$ and ${}^B \mathbf{\Pi}$ are related, we start to write the expression of the point cloud ${}^B \mathbf{\Pi}$ found by using the true homogeneous matrix. It is given by:
 \begin{equation}\label{true_point_cloud}
    {}^B \mathbf{\Pi} =  \bigcup_{j = 1}^{n} {}^B H_{E_j} {}^E H_{E[S]} {}^{E[S]} H_{S} {}^{S_j} \mathbf{P}_j  ,
\end{equation}
it is possible to write the Equation \eqref{true_point_cloud} specifying each term:
\begin{align}\label{explicit_eq_point_cloud}
    {}^B \mathbf{\Pi} = \bigcup_{j = 1}^{n} &
    \begin{bmatrix} {}^{B} R_{E_j} & {}^{B} o_{E_j} \\ 0_{1\times 3} & 1 \end{bmatrix} \nonumber \\
    & \quad
    \begin{bmatrix} {}^{E} R_{S} & 0_{3 \times 1} \\ 0_{1\times 3} & 1 \end{bmatrix}
    \begin{bmatrix} 1_{3 \times 3} & {}^{E[S]} o_{S} \\ 0_{1\times 3} & 1 \end{bmatrix}
    {}^{S_j} \mathbf{P}_j .
\end{align}
The previous equation can be demonstrated to be equal to:
\begin{align}
    {}^B \mathbf{\Pi} = \bigcup_{j = 1}^{n} &
    \Bigg( {}^{B} H_{E_j} {}^{E} H_{E[S]} {}^{S_j} \mathbf{P}_j \nonumber \\
    & \quad
    {+}
    \begin{bmatrix}
        {}^{B} R_{E_j} {}^{E} R_{S} {}^{E[S]} o_{S} \\   0
   \end{bmatrix} \Bigg) .
\end{align}
By noting that ${}^{B} R_{E_j} = {}^{B} R_{E_1}$ (due to Assumption~\ref{ass:eeRotationIsConstant}) and that ${}^E o_S = {}^{E} R_{S} {}^{E[S]} o_{S} $, the term ${}^{B} R_{E_j} {}^{E} R_{S} {}^{E[S]} o_{S}$ can be written as ${}^{B} R_{E_1} {}^{E} o_{S}$, and can be brought outside of the union as it is independent from the $j$ index:
\begin{align}
    {}^B \mathbf{\Pi} = 
   \left( \bigcup_{j = 1}^{n} {}^B H_{E_j} {}^E H_{E[S]} {}^{S_j}\mathbf{P}_j \right) +  \begin{bmatrix} {}^{B} R_{E_1} {}^{E} o_{S}\\   0
   \end{bmatrix}\label{eq:last_step} .
\end{align}
By substituting Equation~\eqref{pointcloudcorrupted} in Equation~\eqref{eq:last_step}, we can conclude that:
\begin{equation}\label{final_eq_pc}
{}^B \mathbf{\Pi} = {}^B \mathbf{\Pi}' + \begin{bmatrix} {}^B R_{E_1} {}^{E} o_{S}\\   0
   \end{bmatrix}  .
\end{equation}

Given Assumption~\ref{ass:eeRotationIsConstant} we assume that we can localize a frame $C$ rigidly attached to a reference object (for example a calibration grid on a plane or a known 3D object) through the point cloud registration process. The 
object is stationary during the point cloud acquisition, such that the homogeneous matrix ${}^B H_C$ does not change. 
With that in mind, it is possible to write a relation based on Equation~\eqref{final_eq_pc} for the origin of the reference frame attached to the object expressed in the robot base frame instead of the full point clouds. We can write:
\begin{equation}
    {}^B o_C = {}^B o_{C'} + {}^B R_{E_1} {}^{E} o_{S} ,
\end{equation}
where ${}^B o_C$ and ${}^{E} o_{S}$ are unknown, ${}^B o_{C'}$ is estimated by the point cloud registration and the end-effector rotation ${}^{B} R_{E_1}$ is known given the robot trajectory.
If the point cloud acquisition is repeated $m$ times we obtain $m$ different reconstructed point clouds and it is possible to build the over-determined system $Ax=b$ with:
\begin{equation}\label{eq:AXBfull}
    A = \begin{bmatrix} 1_3 &  - {}^B R_{E^1_1} \\ 1_3 &  - {}^B R_{E^2_1} \\ \vdots \\ 1_3 &  - {}^B R_{E^m_1} \end{bmatrix} , \;
    x = \begin{bmatrix} {}^B o_C  \\  {}^{E} o_{S} \end{bmatrix} , \;
    b = \begin{bmatrix} {}^B o_{C_1'} \\  {}^B o_{C_2'}  \\  \vdots \\ {}^B o_{C_m'} \end{bmatrix}  .
\end{equation}
The solution $x^*$ to such systems can be found through the least square method, that consists in minimizing the least square error:
\begin{equation}
\label{eq:minlserr}
   x^* = \operatorname{argmin}_x  || Ax - b ||^2  \quad .
\end{equation}
This problem has an analytic solution when the matrix $A$ is full column rank as it turns out that $A^T A$ is invertible, and the unique solution is given by:
\begin{equation}\label{eq:solutionx}
    x^* = (A^T A)^{-1} A^T b  \quad .
\end{equation}

From the solution $x$ of the Equation~\eqref{eq:solutionx} it is possible to obtain the required estimate of the hand-eye translation ${}^{E} o_{S}$.


Equation~\eqref{eq:minlserr} has a unique solution if matrix $A$ in Equation~\eqref{eq:AXBfull} has  full-column rank. By analyzing the nullspace of matrix $A$ it can be shown that the minimum $m$ that respect this condition is $m = 3$, and the rotation matrices ${}^B R_{E^1_1}$ ${}^B R_{E^2_1}$ and ${}^B R_{E^3_1}$ need to respect the following conditions:
\begin{enumerate}
    \item ${}^B R_{E^1_1}$ ${}^B R_{E^2_1}$ and ${}^B R_{E^3_1}$ are all different with respect to each other.
    \item The rotation axis corresponding to the rotation matrices ${}^{E^1_1} R_{E^2_1}$ and ${}^{E^1_1} R_{E^3_1}$ are not parallel.
\end{enumerate}
Even if the minimum number of reconstructed point cloud is $3$, in practice it is convenient to use an higher number of reconstructed point clouds to minimize the effect of the measurement errors in the estimation.


%% file: sections/experiments.tex
\section{Experiments}
\label{sec-experiment}
\begin{figure}[t]
    \centering
    \begin{subfigure}[b]{0.49\columnwidth}
         \centering
         \includegraphics[width=0.6\textwidth]{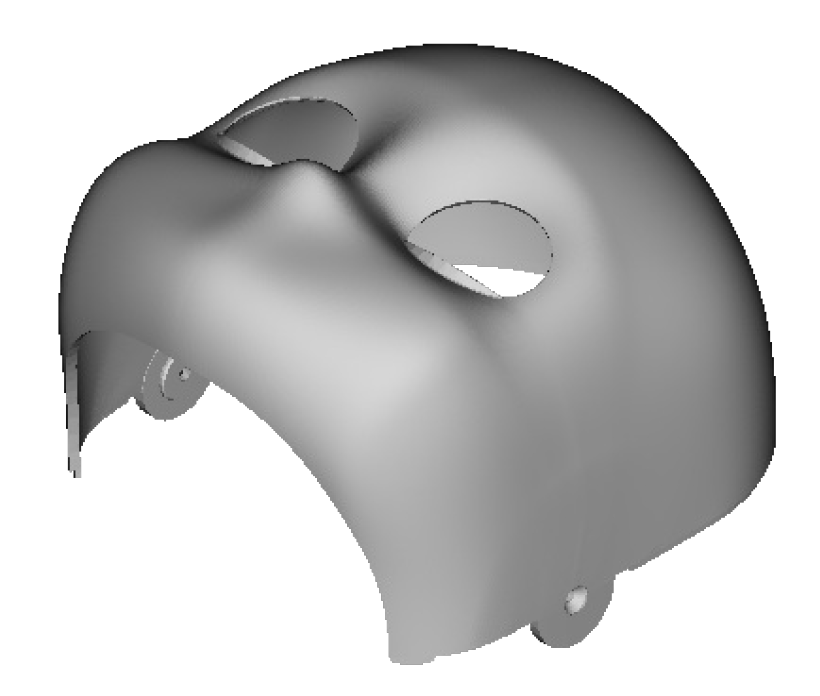}
         \caption{iCub head}
         \label{fig:icub}
     \end{subfigure}
     \hfill
     \begin{subfigure}[b]{0.49\columnwidth}
         \centering
         \includegraphics[width=\textwidth]{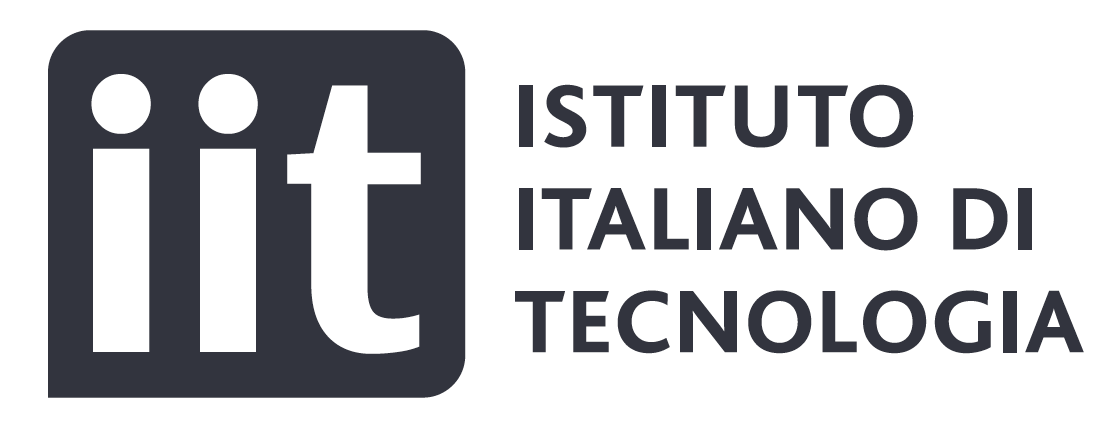}
         \caption{IIT logo}
         \label{fig:logo}
     \end{subfigure}
     \hfill
    \caption{Calibration target: (\ref{fig:icub}) 3D object and (\ref{fig:logo}) 2D image.}
    \label{fig:calib_targets}
\end{figure}
In this section, we present the results of calibration ${}^{E}H_{S}$ using the methodology shown above.
As mentioned earlier, our methodology for the calibration of hand-eye translation is \textit{target-agnostic}, meaning any arbitrary object with known model can be used as a calibration target.
We demonstrate the target-agnostic nature of our approach using two unique and uncommon objects that by itself cannot be considered as calibration targets according to literature. The objects are shown in Figure \ref{fig:calib_targets}.
The Figure \ref{fig:icub} shows the head of the iCub humanoid robot \cite{metta2008icub} which represents the 3D target and Figure \ref{fig:logo} shows the logo of Istituto Italiano di Tecnologia\footnote{https://iit.it/} which represents the 2D target.
In the setup shown in Figure \ref{fig:frames}, a LMI GOCATOR 2380~\cite{gocator_docs} laser profile sensor is mounted onto a KUKA KR30 HA~\cite{kukakr_docs} using a specially designed flange. The system is equipped with the software architecture described in~\cite{rendiniello2020flexible} to synchronize the laser and robot data.
For the LMI GOCATOR 2380, the Z linearity is 0.32 mm and the Z repeatability of the  is $\pm$12$\mu$m ~\cite{gocator_docs}. The pose repeatability of the KUKA KR30 HA is $\pm$ 0.05 mm~\cite{kukakr_docs}.

The calibration target is placed on a workbench in the workspace of the robot to allow for multiple dataset acquisitions and the vertical distance between the target and the sensor is ensured to be within the near and far field-of-view (FoV) of the sensor. 
The unit for the coordinate system is millimeter (mm) and degrees (deg). 
The work-flow of the experiment is data acquisition, point cloud registration and calibration. The workflow is identical for both 2D and 3D targets unless otherwise specified as detailed below.

\subsection{Data acquisition}\label{subsec:data_acquisition}

We collected $5$ datasets for the 2D target and 3D target respectively where each dataset is composed of $10$ point clouds. Each point cloud is reconstructed starting from laser profiles acquired by moving the robot from a starting point to a final one keeping the orientation constant to respect the Assumption~\ref{ass:eeRotationIsConstant} of Section~\ref{sec:methods}.
Furthermore, we ensure that between different reconstructed point clouds, the orientation of the end-effector ${}^B R_{E_1^{i}}$ is varied while keeping the calibration target stationary and within the FoV.
Referring to Figure \ref{fig:frames}, the user provides \textit{start pose}, \textit{end pose} and number of steps such that the calibration target is scanned completely by the laser. The data acquisition program performs linear interpolation for the given number of steps between the given poses and sends the commands to the robot following the \textit{stop and look} procedure explained in Section \ref{sec:introdcution}. More formally, given $\{X_{start}, X_{end}, N\}$ where $N$ is the total number of steps and $X = \{ x,y,z \}$ the program computes
\begin{equation}
    X_i = X_{start} + r/N \times  (X_{end} - X_{start})   ,
\end{equation}
for $r \in \{0, 1 \dots N\}$.
As the laser profile sensors provide both depth and intensity data ($x^S, z^S, I^S$ respectively), the extracted profiles are transformed into the robot base frame as explained in Section~\ref{sec:methods} and encoded as a point cloud data structure consisting of the Euclidean XYZ coordinates and the grayscale intensity value \texttt{pcl::PointXYZI}~\cite{Rusu_ICRA2011_PCL}. 
The use of intensity data in addition to the Euclidean XYZ coordinates is to assist point cloud segmentation for 2D targets. In case of a 3D target, the methodology is identical using only the Euclidean XYZ coordinates.
The reconstructed point cloud extracted using the laser scans is henceforth referred as the \textit{scene point cloud}.
In order to perform point cloud registration, we also require a \textit{model point cloud}. For 3D targets, the model point cloud is provided by mesh sampling the given 3D CAD model of the target. 

For 2D targets i.e., an image, we use the following relation to form a model point cloud:
\begin{equation}
\begin{split}
   x &= u \times \mathrm{mm/Pixel}, \\ 
   y &= v \times \mathrm{mm/Pixel}, \\ 
   z &= 1,\\            
   I &= \mathrm{pixelIntensity(u,v)},
\end{split}
\end{equation}
for each pixel $(u,v)$ in the image. $x,y,z,I$ form the coordinates of the corresponding pixel in the point cloud. The $\mathrm{mm/Pixel}$ parameter is derived from dots-per-inch (DPI) of the image. The program iterates through each pixel to extract its position and intensity by using OpenCV functions~\cite{opencvImage}.

\subsection{Point cloud registration}
Point cloud registration is the process of finding a spatial transform that aligns two point clouds. Prior to registration, the scene clouds require pre-processing in order to segment and remove noisy data.
In the case of 2D targets, we have strong dark and light regions in the \textit{scene} and \textit{model}, we perform \textit{binarization} to choose only the points having high intensity. The binarization can be done by simply choosing a proper threshold for the intensity and removing all points which fall below the intensity threshold. This pre-processing step also effectively removes spurious outliers that can affect the quality of point cloud registration. In the case of 3D targets, we pre-process the scene cloud using Statistical Outlier Removal filter that removes sparse and noisy points. The point cloud registration is done using RANSAC~\cite{fischler1981random} to provide a good initial alignment and by ICP~\cite{besl1992method} to perform the fine registration. 

\subsection{Calibration}
The point cloud registration for each reconstructed point cloud for the 2D and 3D target provides ${}^{B} H_{C}$ of which we are interested in ${}^B o_{C}$ where $C$ is the frame rigidly attached to the target. Plugging in the values ${}^B o_{C_i'}$ for $i \in 1, \dots, m$ (where $m$ is the number of reconstructed point clouds) in vector $b$ and the values of  ${}^B R_{E^i_1}$ for matrix $A$ from  we can solve Equation \eqref{eq:AXBfull} for $x$ as 
\begin{equation}
    x = A^{+}b,
\end{equation}  
where ${}^{+}$ is the pseudo-inverse of the matrix. There are various methods of solving the pseudo-inverse problem and in particular we use the Householder rank-revealing QR decomposition of a matrix with column-pivoting of C++ Eigen library \cite{psuedoInv}.

\subsection{Results}
\begin{figure}[t]
    \centering
    \begin{subfigure}[b]{0.235\textwidth}
         \centering
         \includegraphics[width=\textwidth, height = 3cm]{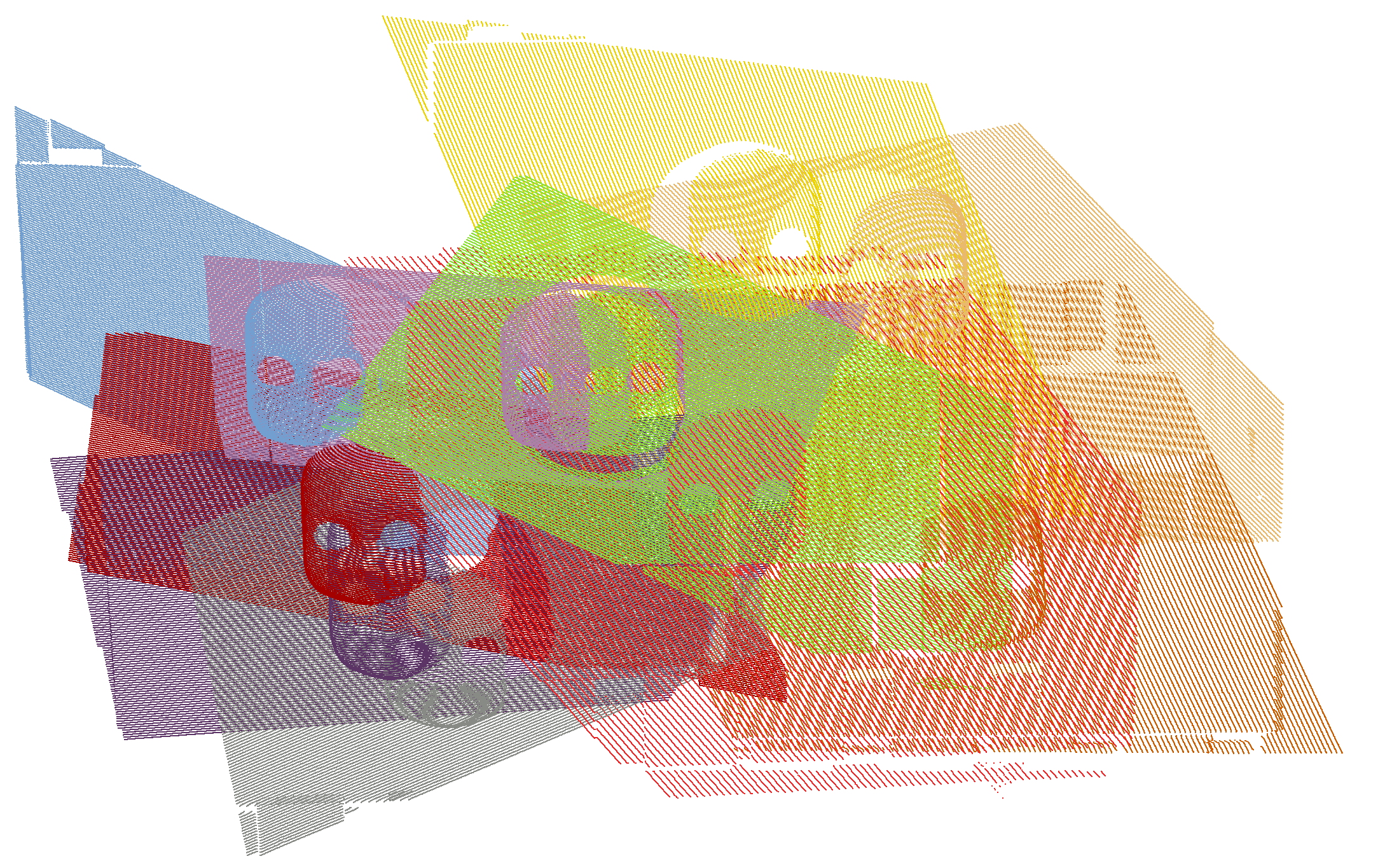}
         \caption{Incorrect hand-eye translation calibration}
         \label{fig:3duncalibrated}
     \end{subfigure}
     \begin{subfigure}[b]{0.235\textwidth}
         \centering
         \includegraphics[width=\textwidth, height = 3cm]{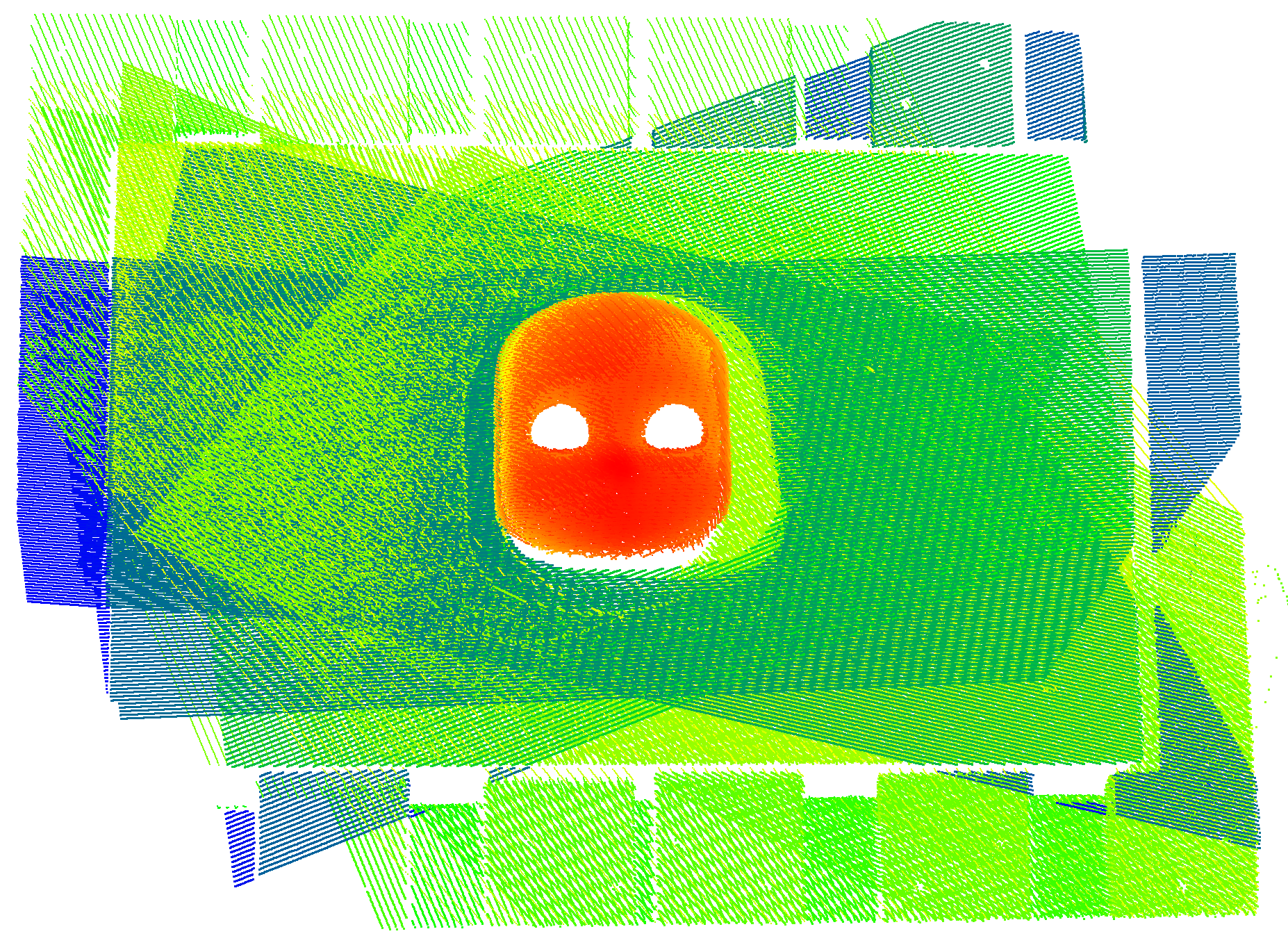}
         \caption{Correct hand-eye translation calibration}
         \label{fig:3dcalibrated}
     \end{subfigure}
     \hfill
    \caption{The effect of incorrect translation (\ref{fig:3duncalibrated}) and correct translation (\ref{fig:3dcalibrated}) in the hand-eye calibration matrix ${}^E H_{S}$ of the stationary 3D target with 10 different scanning trajectories.}
    \label{fig:3dcalibration}
\end{figure}
\begin{figure}[t]
    \centering
    \begin{subfigure}[b]{0.235\textwidth}
         \centering        \includegraphics[width=\textwidth, height = 3cm]{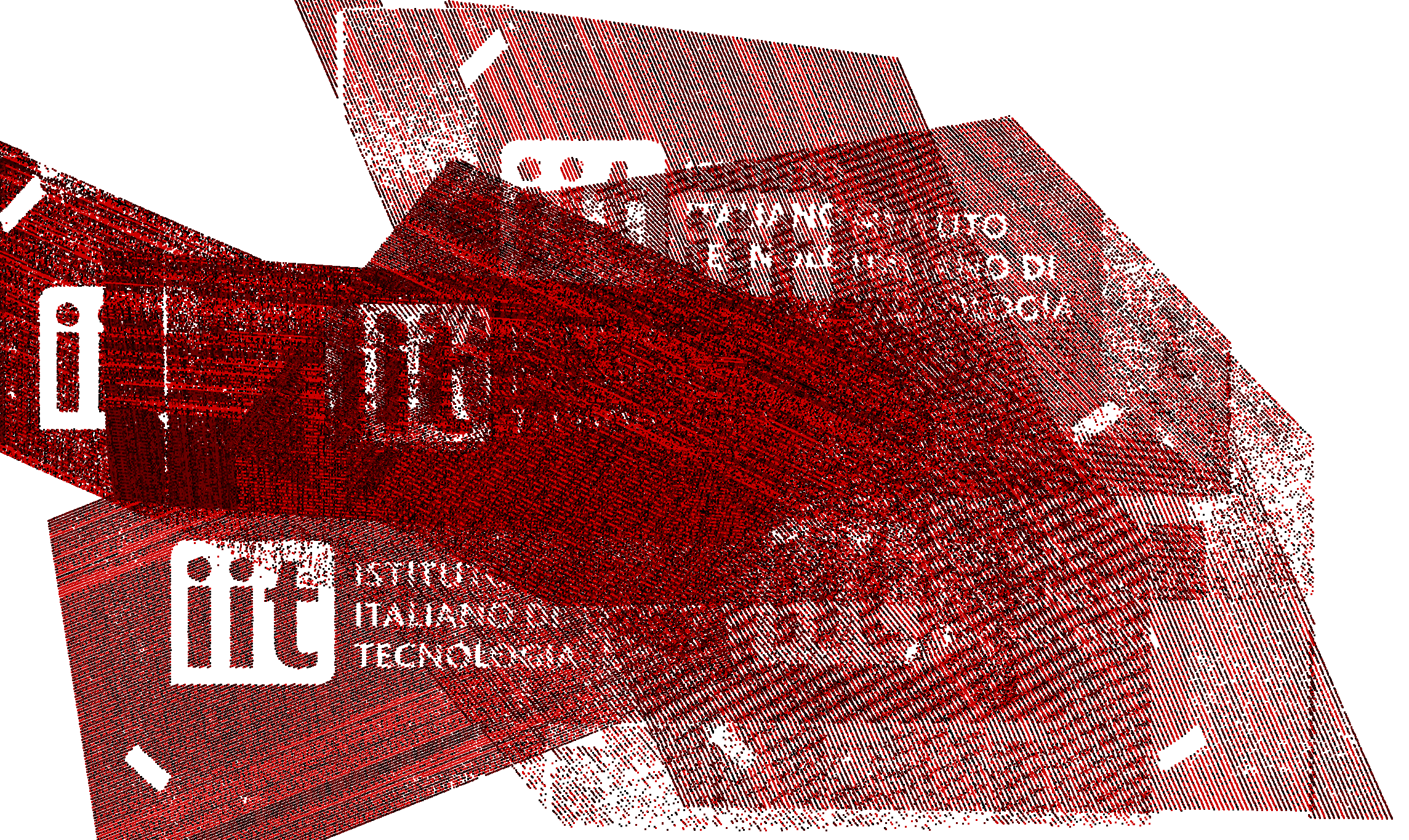}
         \caption{Incorrect hand-eye translation calibration}
         \label{fig:2duncalibratedlogo}
     \end{subfigure}
     \hfill
     \begin{subfigure}[b]{0.235\textwidth}
         \centering
         \includegraphics[width=\textwidth, height = 3cm]{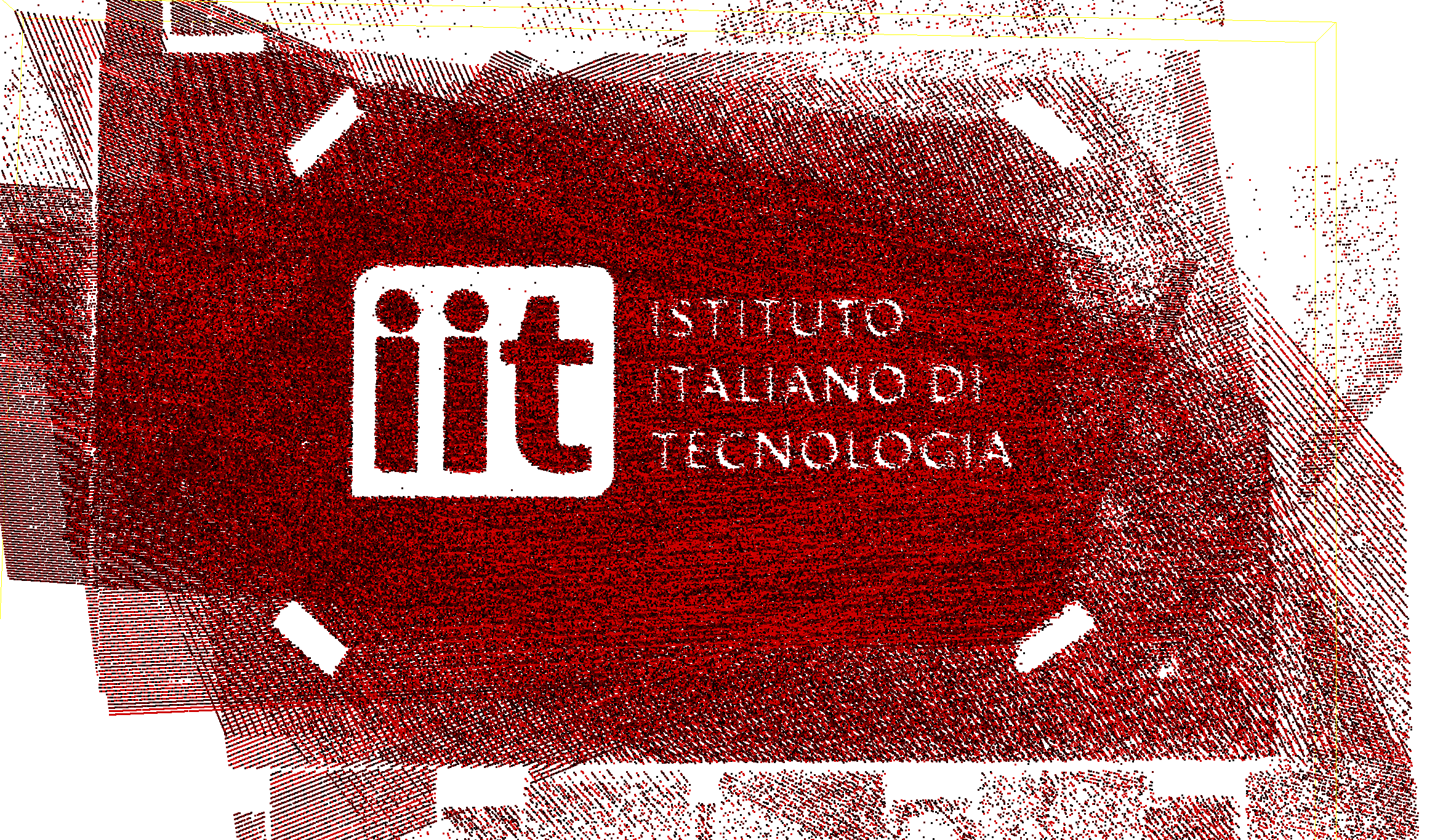}
         \caption{Correct hand-eye translation calibration}
         \label{fig:2dcalibratedlogo}
     \end{subfigure}
     \hfill
    \caption{The effect of incorrect translation (\ref{fig:2duncalibratedlogo}) and correct translation (\ref{fig:2dcalibratedlogo}) in the hand-eye calibration matrix ${}^E H_{S}$ of the stationary 2D target with 10 different scanning trajectories.}
    \label{fig:2dcalibration}
\end{figure}
In order to evaluate the results of the calibration process, we extract the ground-truth transformation from the CAD model of the sensor-holding flange and of the sensor provided by the vendor. The ground-truth translation vector extracted in this way is $\{907.5~\mathrm{mm},~ 97.0~\mathrm{mm},~40.0~\mathrm{mm}\}$. As aforementioned, the rotation part ${}^{E} R_{S}$ in \eqref{eq:frame_with_only_rotation_known} is known \textit{a priori} and is determined from the mechanics of the flange support as:
\begin{equation}\label{rotation-value}
    {}^{E} {R}_{S} = \begin{bmatrix}
        0.0 &  0.0 & -1.0 \\
        0.0 &  -1.0 &  0.0 \\
       -1.0 &  0.0 &  0.0 \\
    \end{bmatrix}   .
\end{equation}
The result of the calibration is the translation vector $\{x, y, z \}$ and the corresponding error with respect to the ground truth translation $\{\Delta x, \Delta y, \Delta z \}$ is presented in Table \ref{tab:2dresult} (2D target) and Table \ref{tab:3dresult} (3D target) for the $5$ acquired datasets. Furthermore, Figure \ref{fig:3dcalibration} and \ref{fig:2dcalibration} shows the effect of incorrect and correct hand-eye translation calibration for 
one dataset overlayed for a stationary 3D and 2D target respectively.

\begin{figure}[t]
    \centering
    \begin{subfigure}[b]{\columnwidth}
         \centering        
         \includegraphics[width=0.8\textwidth, height = 4cm]{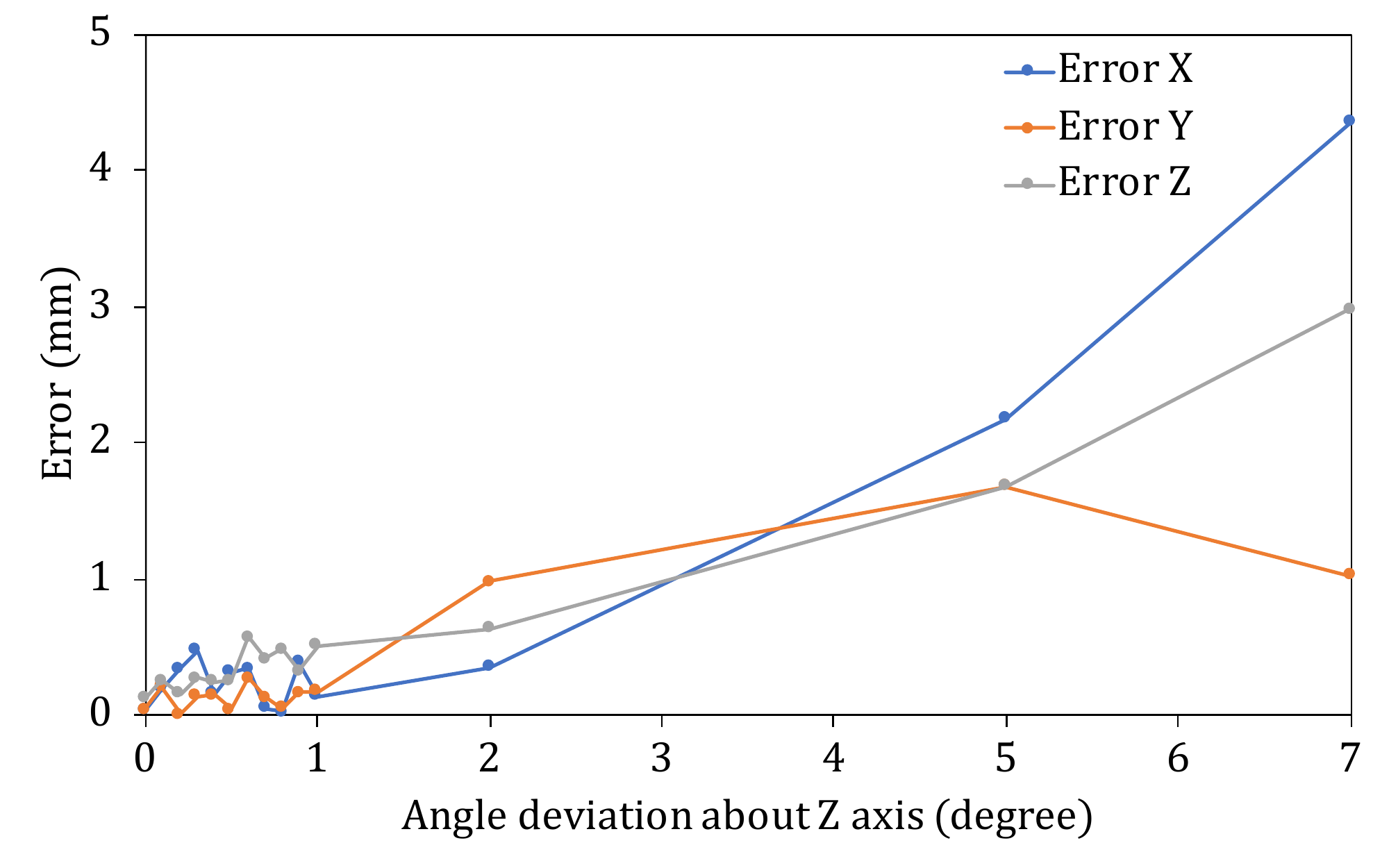}
         \caption{Drift in calibration with perturbation of known rotation component.}
         \label{fig:angledev}
     \end{subfigure}
     \hfill
     \begin{subfigure}[b]{\columnwidth}
         \centering
         \includegraphics[width=0.84\textwidth, height = 4cm]{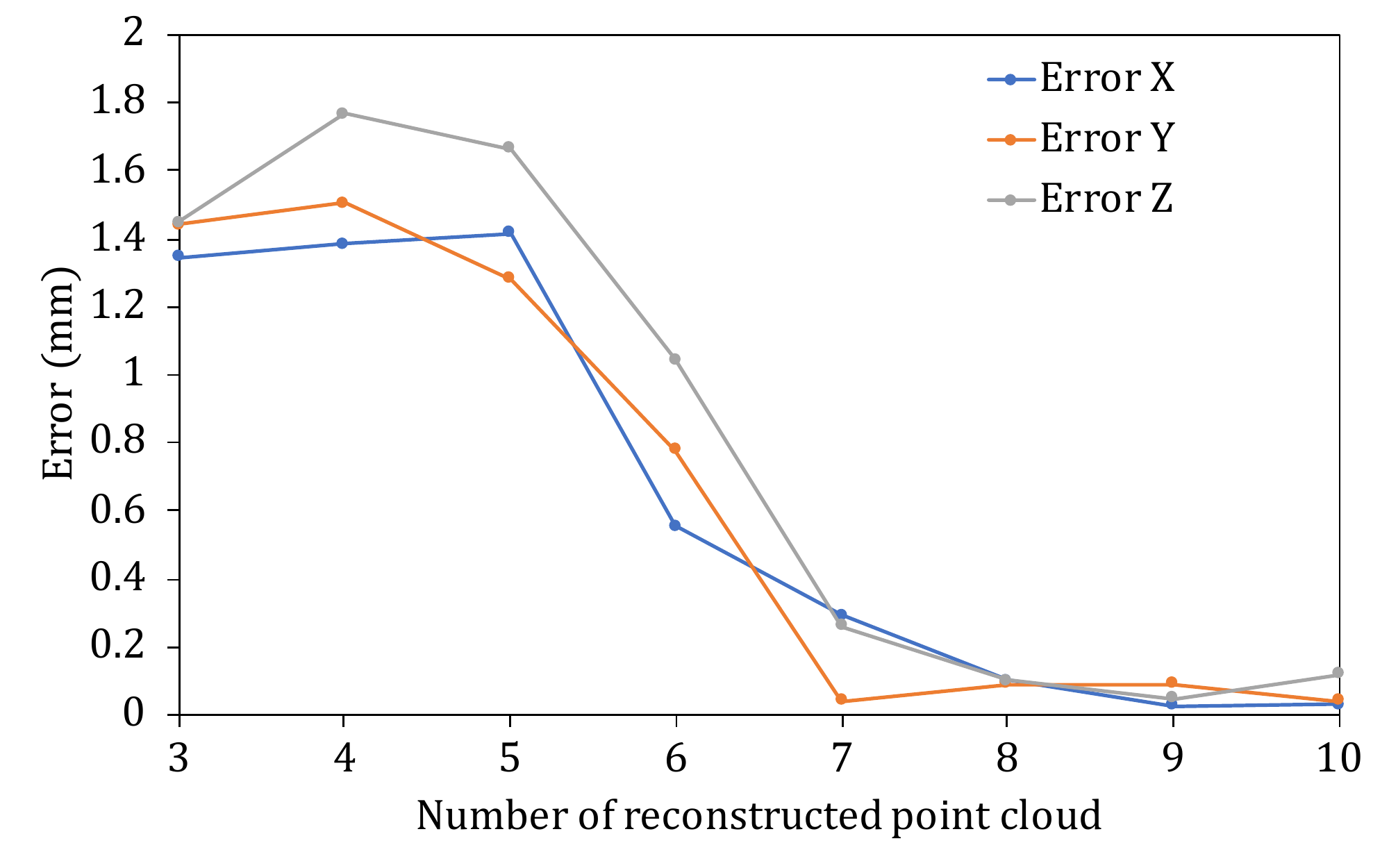}
         \caption{Drift in calibration with the number of reconstructed point clouds.}
         \label{fig:minimum_clouds}
     \end{subfigure}
     \hfill
    \caption{(\ref{fig:angledev}) The drift in calibration with  perturbation in prior known rotation. The perturbation   is provided along the $z$ axis for ${}^{E} R_{S}$. (\ref{fig:minimum_clouds}) The drift in calibration with the number of reconstructed point clouds composing the dataset used for calibration.}
    \label{fig:stressTest}
\end{figure}

The obtained results for both experiments show that the calibration results have sub-millimeter accuracy and precision. In particular, it is possible to note that there is relatively less consistency with respect to the ground-truth in the results obtained with the 3D target than the one obtained with the 2D target. This can be explained by the trajectory-specific registration errors that are higher for the 3D target, given its specific shape for which different parts of the object can be occluded depending on the trajectory. For the 2D target instead, the entire image is  visible for all trajectories, minimizing the sources of registration errors.
Furthermore, in order to investigate the accuracy of the calibration given noisy prior knowledge of the hand-eye rotation, we provided an incremental perturbation about the Z axis for ${}^{E} R_{S}$ i.e., $\Delta R_z$ for the 2D target and performed the calibration using the collected dataset consisting of 10 reconstructed point clouds. The results shown in Figure \ref{fig:angledev} demonstrates that for minor perturbations up to $2^{o}$, the errors ($\Delta x$, $\Delta y$, $\Delta z$) are $\leq 1~mm$ while the errors increase with increasing perturbation. While our proposed method can handle small perturbations ($1^{o} \sim 2^{o}$), it is important to have an accurate ground truth hand-eye rotation as defined in Assumption~\ref{ass:rotationIsKnown} in order to extract the correct hand-eye translation. On the other hand, Figure~\ref{fig:minimum_clouds} shows the variation in accuracy of hand-eye translation calibration with the number of reconstructed point clouds ranging from 3 to 10. As the number of reconstructed point clouds composing the dataset increases, the accuracy of calibration also improves. In the particular dataset used for computing Figure~\ref{fig:minimum_clouds} and given ordering of the reconstructed point clouds, the theoretic minimum number of reconstructed point clouds i.e., three point clouds provides calibration with mean error of $1.41 ~mm$ whereas the mean error reduces to $0.196 ~mm$ with seven point clouds.

\begin{table}[t!]
\begin{center}
\resizebox{\columnwidth}{!}{%
\begin{tabular}{|c||c|c|c|c|c|c|}
\hline
\textbf{Dataset} & \textbf{x (mm)} & \textbf{y (mm)} & \textbf{z (mm)} & \textbf{$\Delta$x (mm)} & \textbf{$\Delta$y (mm)} & \textbf{$\Delta$z (mm)} \\ \hline \hline 
\textbf{1}       & 907.472    & 96.961     & 39.882     & 0.028          & 0.039          & 0.117          \\ \hline
\textbf{2}       & 907.276    & 97.019     & 39.938     & 0.224          & 0.018          & 0.062          \\ \hline
\textbf{3}       & 907.116    & 97.011     & 39.995     & 0.384          & 0.011          & 0.005          \\ \hline
\textbf{4}       & 907.177    & 96.959     & 39.945     & 0.323          & 0.041          & 0.054          \\ \hline
\textbf{5}       & 907.188    & 97.121     & 39.990     & 0.312          & 0.121          & 0.009          \\ \hline
\textbf{Mean} & 907.246    & 97.014     & 39.950     & 0.254          & 0.046          & 0.049          \\ \hline
\textbf{SD}     & 0.139     & 0.066      & 0.046      & 0.139          & 0.044          & 0.046          \\ \hline
\end{tabular}}%
\end{center}
\caption{Calibration result for 10 sets of acquisition repeated 5 times for the 2D target. SD -- Standard deviation}
\label{tab:2dresult}
\end{table}

\begin{table}[t!]
\begin{center}
\centering
\resizebox{\columnwidth}{!}{%
\begin{tabular}{|c||c|c|c|c|c|c|}
\hline
\textbf{Dataset} & \textbf{x (mm)} & \textbf{y (mm)} & \textbf{z (mm)} & \textbf{$\Delta$x (mm)} & \textbf{$\Delta$y (mm)} & \textbf{$\Delta$z (mm)} \\ \hline \hline 
\textbf{1}       & 906.881    & 96.653     & 40.384     & 0.619          & 0.347          & 0.384          \\ \hline
\textbf{2}       & 906.949    & 96.611     & 40.434     & 0.551          & 0.389          & 0.434          \\ \hline
\textbf{3}       & 906.766    & 96.416     & 40.283     & 0.734          & 0.584          & 0.283          \\ \hline
\textbf{4}       & 906.592    & 96.469     & 40.394     & 0.908          & 0.531          & 0.394          \\ \hline
\textbf{5}       & 906.810    & 96.638     & 40.337     & 0.690          & 0.362          & 0.337          \\ \hline
\textbf{Mean}    & 906.797    & 96.557     & 40.366     & 0.701          & 0.443          & 0.366          \\ \hline
\textbf{SD}     & 0.135      & 0.107      & 0.058      & 0.135          & 0.107          & 0.058         \\ \hline
\end{tabular}}%
\end{center}
\caption{Calibration result for 10 sets of acquisition repeated 5 times for the 3D target. SD -- Standard deviation}
\label{tab:3dresult}
\end{table}



%% file: sections/conclusions.tex
\section{Conclusions}
\label{sec:conclusions}

We present a novel method for translation-only hand-eye calibration of a 2D laser sensor on a robotic manipulator. With the assumption of known rotation for the hand-eye calibration matrix and estimated positions of the calibration target through point cloud registration, the problem of finding the translation hand-eye calibration boils down to solving an over-determined system of linear equations of the form $Ax=b$. The method is demonstrated to be easy to implement for practitioners, robust and highly accurate. We implement our method on a real robotic platform and validate its repeatability with statistical analysis. We also demonstrate that our proposed method is robust to minor perturbations (up to $2^{o}$) in the prior knowledge of hand-eye rotation.
The novelty in this paper lies in the \textit{target-agnostic} method wherein the approach is applicable to any arbitrary 2D or 3D calibration object and can also be performed \textit{in situ} in the final application. 
The limitation of this work lies in the assumption that hand-eye rotation is known \textit{a priori} (Assumption~\ref{ass:rotationIsKnown}). 
This is considered as future work wherein the complete computation (translation and rotation) of hand-eye calibration matrix via \textit{any} arbitrary calibration target can greatly simplify the tedious calibration process in industrial applications. Another direction for future work can be the generation of optimal trajectories to reconstruct the given target for accurate hand-eye calibration.